\documentclass[acmtog,nonacm]{acmart}
 
\copyrightyear{2024} 
\acmYear{2024} 
\setcopyright{rightsretained} 

\usepackage{multirow}
\usepackage{tabularx}
\newcolumntype{Y}{>{\centering\arraybackslash}X}
\usepackage{booktabs}
\usepackage{colortbl}
\usepackage{color}
\definecolor{niceblue}{rgb}{0, 0.635, 0.929}

\graphicspath{ {./figures} }

\newcommand{\fairfacecopy}{\footnotesize \textit{FairFace images -- CC BY 4.0.}}

\begin{document}

\title{Hairmony: Fairness-aware hairstyle classification}


\author{Givi Meishvili}
\email{gmeishvili@microsoft.com}
\orcid{0000-0002-0984-7078}
\authornote{denotes equal contribution}
\author{James Clemoes}
\email{jaclemoe@microsoft.com}
\orcid{0009-0009-5083-201X}
\authornotemark[1]
\author{Charlie Hewitt}
\email{chewitt@microsoft.com}
\orcid{0000-0003-3943-6015}
\authornotemark[1]
\author{Zafiirah Hosenie}
\email{zhosenie@microsoft.com}
\orcid{0000-0001-6534-593X}
\author{Xian Xiao}
\email{xianxiao@microsoft.com}
\orcid{xianxiao@microsoft.com}
\author{Martin de La Gorce}
\email{martin.delagorce@microsoft.com}
\orcid{0009-0007-3739-1980}
\author{Tibor Takacs}
\email{Tibor.Takacs@microsoft.com}
\orcid{0009-0002-4185-4654}
\author{Tadas Baltru\v{s}aitis}
\email{tabaltru@microsoft.com}
\orcid{0000-0001-7923-8780}
\author{Antonio Criminisi}
\email{acriminisi@microsoft.com}
\orcid{0000-0002-3668-7014}
\author{Chyna McRae}
\email{chynamcrae@microsoft.com}
\orcid{0009-0000-1907-944X}
\affiliation{\institution{Microsoft}\city{Redmond}\country{Switzerland, United Kingdom and United States of America}}
\author{Nina Jablonski}
\email{ngj2@psu.edu}
\orcid{0000-0001-7644-874X}
\affiliation{\institution{The Pennsylvania State University}\city{University Park}\country{United States of America}}
\author{Marta Wilczkowiak}
\email{mawilczk@microsoft.com}
\orcid{0009-0006-7695-4216}
\affiliation{\institution{Microsoft}\country{United Kingdom}}

\renewcommand{\shortauthors}{Meishvili et al.}

\begin{abstract}
We present a method for prediction of a person's hairstyle from a single image.
Despite growing use cases in user digitization and enrollment for virtual experiences, available methods are limited, particularly in the range of hairstyles they can capture.
Human hair is extremely diverse and lacks any universally accepted description or categorization, making this a challenging task.
Most current methods rely on parametric models of hair at a strand level.
These approaches, while very promising, are not yet able to represent short, frizzy, coily hair and gathered hairstyles.
We instead choose a classification approach which \emph{can} represent the diversity of hairstyles required for a truly robust and inclusive system.
Previous classification approaches have been restricted by poorly labeled data that lacks diversity, imposing constraints on the usefulness of any resulting enrollment system.
We use only synthetic data to train our models. This allows for explicit control of diversity of hairstyle attributes, hair colors, facial appearance, poses, environments and other parameters. It also produces noise-free ground-truth labels.
We introduce a novel hairstyle taxonomy developed in collaboration with a diverse group of domain experts which we use to balance our training data, supervise our model, and directly measure fairness.
We annotate our synthetic training data and a real evaluation dataset using this taxonomy and release both to enable comparison of future hairstyle prediction approaches.
We employ an architecture based on a pre-trained feature extraction network in order to improve generalization of our method to real data and predict taxonomy attributes as an auxiliary task to improve accuracy.
Results show our method to be significantly more robust for challenging hairstyles than recent parametric approaches.
Evaluation with taxonomy-based metrics also demonstrates the fairness of our method across diverse hairstyles.

\end{abstract}



\begin{teaserfigure}
    \centering
    \includegraphics[width=\linewidth]{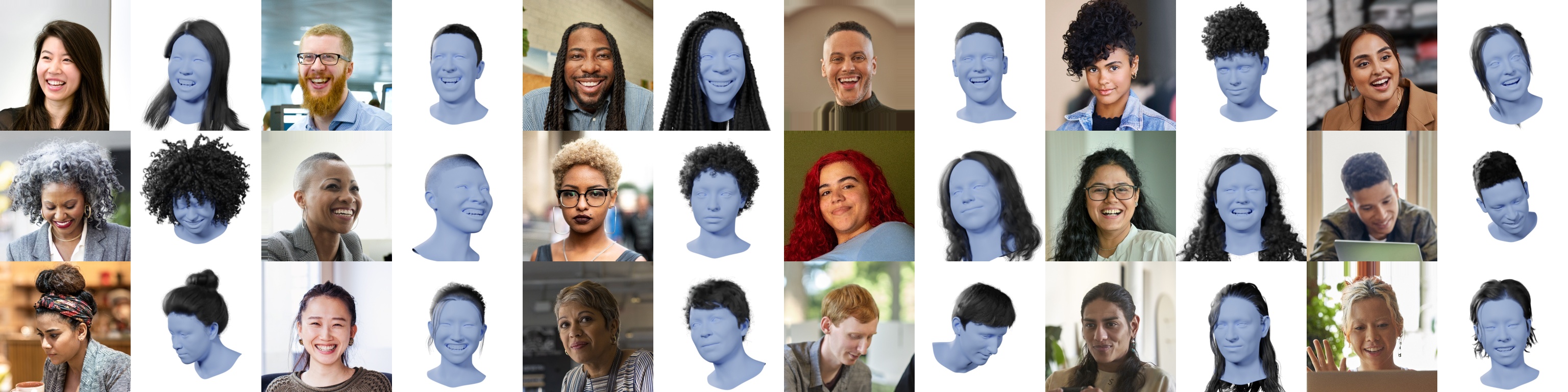}
    \caption{
    Our method predicts a person's hairstyle from a single input image. We use a classification approach and a carefully designed hairstyle library, which allows us to address a wide range of hairstyles, including those that are still an open area of research for parametric models, such as ringlets, braids, and ponytails. 
    Here we show input images and predicted hairstyles from our library posed on reconstructed face meshes~\cite{wood20223d}.
    Underpinning our method is a hairstyle taxonomy that enables objective description of hairstyles to help ensure fairness of our training data and our hairstyle library, supervise our model during training, and evaluate our method.
    }
    \label{fig:teaser}
\end{teaserfigure}

\maketitle

\section{Introduction}
We address the problem of hairstyle prediction from a single image.
This is particularly relevant for enrollment of user appearance for virtual experiences.
It is critical for enrollment systems in this context to work well for diverse users in order to provide an inclusive experience.
Hairstyles play a significant role in human appearance and identity, allowing for self-expression, artistic expression, and cultural representation. 
Inclusive hair representation is also protected by law in many places, with legal guidelines against race discrimination based on hair texture and style \cite{nychrl, crown, ehrc}.
Example results of our method are shown in \autoref{fig:teaser}.

Despite its importance, accurately predicting hairstyles remains an ongoing challenge. 
This is primarily due to the complex and intertwined factors associated with hairstyles, including hair density, thickness, curliness, length, growth direction, melanin composition, texture, and styling. 
Reflecting this complexity is the absence of a universally accepted description, whether linguistic or mathematical, that adequately captures the diversity of human hairstyles.

Existing hair prediction methods are split into two groups: hairstyle classification or attribute prediction; and direct parametric modeling of hair strands.
Classification or attribute based systems have so far been severely limited by available descriptions of hair.
Hairstyle names are culturally specific \cite{encyclopediahair, gittens2002african, corson2001fashions} and their attributes are often described using subjective, inconsistent and ambiguous terms \cite{lasisi2021constraints, shapevariability, wu2023logical}. 
Poor categorization or description has therefore led to systems which do not support the full range of human hairstyles.
Data has also been a significant limitation as few datasets are available and annotations are often of poor quality.
Parametric approaches have also been limited by data, both lacking volume and diversity.
More fundamentally, however, these methods often are simply not formulated to account for frizzy, coily or gathered styles where high-frequency variations in strand shapes are common. 

We choose a classification approach to avoid the limitations inherent to current strand-based methods, and aim to address the shortcomings of prior work by improving both on the fair description and categorization of hair through a novel hairstyle taxonomy, and the quality of data by using synthetic training images.
In so doing we prioritize fairness and robustness over accuracy and directness of representation.
We acknowledge the scalability of direct, parametric approaches is desirable; however we believe our approach 
is an important step towards inclusive hairstyle prediction systems.
Specifically, our contributions are as follows:
\begin{itemize}
    \item A novel hairstyle taxonomy developed in collaboration with diverse domain experts and iteratively refined through pilot studies. 
    \item Synthetic training and real evaluation datasets for hairstyle prediction annotated with our taxonomy, which we release to facilitate future work.
    \item A hairstyle prediction method from single images using taxonomic losses that generalizes well to real test data, despite being trained only on synthetic images.
\end{itemize}
The full taxonomy definition and dataset download instructions can be found on the project website at {\color{niceblue}\url{https://aka.ms/hairmony}}.

\section{Related Work}
\autoref{tab:hair_type_coverage} provides a summary of recent work in the space and its coverage of different hair types and features critical for inclusive hair description/prediction.
The rest of this section gives an overview of these works and others, and how our method compares to them.

\begin{table}
    \caption{Hair diversity support in existing work: While 3D hairstyle regression techniques are making rapid progress in extending their support to more complex hairstyles, support for coily hair, strands (such as braids and dreadlocks) and gathering (such as ponytails and buns) remains an open problem. 
\label{tab:hair_type_coverage}
}
    \centering
    \resizebox{\linewidth}{!}{%
    \footnotesize
    \begin{tabular}{@{}lccccccc@{}}
         \toprule
         & \includegraphics[width=1.5cm]{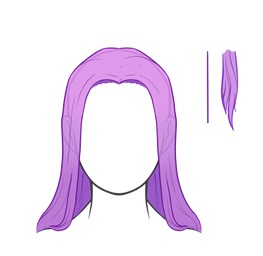} &
         \includegraphics[width=1.5cm]{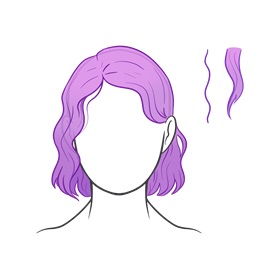} &
         \includegraphics[width=1.5cm]{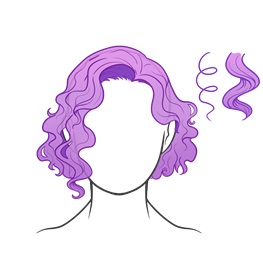} &
         \includegraphics[width=1.5cm]{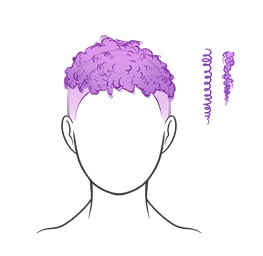} & 
         \includegraphics[width=1.5cm]{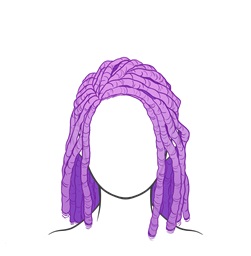} &
         \includegraphics[width=1.5cm]{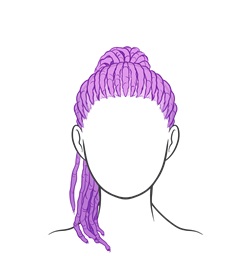} &
         \includegraphics[width=1.5cm]{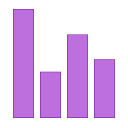} \\
         & \multirow{2}*{Straight} & \multirow{2}*{Wavy} & \multirow{2}*{Curly} & \multirow{2}*{Coily} & Braids/ & Ponytails/ & Fairness \\
         &  &  &  &  & Dreadlocks & Gathered & Quantified \\
         \midrule
         \citet{hu2017avatar} & \checkmark & \checkmark \\
         \citet{sklyarova2023neural} & \checkmark & \checkmark \\
         \citet{yang2019dynamic} & \checkmark & \checkmark \\
         \citet{deep_mvs_hair} & \checkmark & \checkmark & \checkmark \\
         \citet{Wu_2022_CVPR} & \checkmark & \checkmark & \checkmark \\
         \citet{Zheng_2023_CVPR} & \checkmark & \checkmark & \checkmark \\
         \citet{kim2021k} & \checkmark & \checkmark & \checkmark\\
         \citet{rosu2022neural} & \checkmark & \checkmark & & & & \checkmark\\
         \citet{zhou2023groomgen} & \checkmark & \checkmark & \checkmark & \checkmark \\
         \citet{sklyarova2023haar} & \checkmark & \checkmark & \checkmark & \checkmark \\
         \citet{shen2023CT2Hair} & \checkmark & \checkmark & \checkmark & \checkmark \\
         \citet{chen2021celebhair} & \checkmark & \checkmark & \checkmark & \checkmark \\
         \citet{yin2017learning} & \checkmark & \checkmark & \checkmark & \checkmark & \checkmark & \checkmark \\
         \textbf{Ours} & \checkmark & \checkmark & \checkmark & \checkmark & \checkmark & \checkmark & \checkmark \\
         \bottomrule
    \end{tabular}%
    }
\end{table}

\subsection{Description and Classification of Hairstyle}
\label{sec:past_work_class}

Objective description of hairstyle is a very challenging problem given the vast range of hair types and styles present across humans.
Hair texture classification systems describe natural attributes of hair ~\cite{Curlcentric} and are primarily driven by the beauty industry. 
\citet{andrewalker} assigns values 1-4 to hair with varying degrees of curliness.
\citet{shapevariability} introduce 8 classes based on statistical analysis of physical measurements of hair strands from subjects of diverse ancestry.
These systems describe only physical characteristics of hair strands, and not styling, so are useful but not sufficient for the task of hairstyle prediction.

There are two main methods for describing head hair \emph{styling}: using attributes referring to a particular aspect of the style (e.g., bangs type, length, updo type) or using the name of an entire style (e.g., bob, Bantu knots, man bun) \cite{yin2017learning, hu2017avatar, svanera2016figaro}. 
\citet{chen2021celebhair} used a large-scale dataset, CelebHair, to train a model for hairstyle prediction. \citet{kim2021k} introduced a Korean hairstyle dataset and validated its effectiveness via several applications, including hairstyle classification.
\citet{wu2023logical} proposed a logically consistent hair attribute prediction loss to impose consistency across attributes.

As observed by \citet{wu2023logical}, existing hairstyle datasets are limited in consistency and completeness of annotations. 
Specifically: \emph{(1) inconsistency and ambiguity}, \citet{kim2021k} annotate hair length as either short, medium, long, male or female;
\emph{(2) subjectivity}, understanding the attributes requires a specific cultural background (``Rachel'', ``bouffant'') and may be considered culturally inappropriate (``kinky'', ``shag''); 
\emph{(3) incompleteness}, datasets annotated with hairstyles (as opposed to attributes) are limited to specific parts of the population, e.g., \citet{yin2017learning} focus on straight and curly styles, with only 4 out of 64 styles 
representing coily hair and \citet{kim2021k} focus on Korean hair. 

We introduce a taxonomy to describe both hair \emph{type} and \emph{styling} that is designed to be both objective and inclusive, with no emphasis on any particular demographic.
The aim of the taxonomy is to be complete enough to capture all human hairstyles with reasonable accuracy, and granular enough to distinguish between the majority of styles.
We recognize that achieving a fully comprehensive taxonomy is challenging, so aim for simplicity and extensibility to accommodate additional hair attributes and future hairstyles as required.
We choose to perform hairstyle classification, ensuring robust and fair results using our taxonomy to inform improvements to our hairstyle library and evaluate our trained models.
We also predict taxonomy attributes as an auxiliary task to improve hairstyle prediction performance.
We design the taxonomy to be logically consistent, so do not need to enforce consistency through loss functions~\cite{wu2023logical}.



\subsection{Hairstyle Reconstruction} 
\label{sec:param_hair}

Other recent methods choose to \emph{directly} represent hair strips or strands parametrically.
\citet{hu2017avatar} proposed a polygonal strip representation and a deep neural networks (DNN) for semantic hair attribute classification.
\citet{Yuksel_2009} and \citet{jung_2018} used  hair meshes to generate individual hair strands for a wide range of realistic hairstyles.
\citet{zhou2018hairnet} introduced a collision loss to generate full 3D hair geometry while \citet{saito20183d} used a volumetric parameterization to produce a compact latent space of 3D hairstyles.
These methods work well for long straight and wavy hair (with texture captured using polygonal strips), but struggle with the high frequency components typical of frizzy and coily hair.

More recently, neural scalp textures have become popular for hair representation \cite{rosu2022neural,deep_mvs_hair}.
\citet{zhou2023groomgen} combined neural scalp textures with a Fourier-space strand parameterization to better capture high frequency components of strand shape, enabling more accurate representation of frizzy hair, and \citet{sklyarova2023haar} combined neural scalp textures with a text embedding to generate 3D hairstyles, demonstrating some results for frizzy hair. Both works only model hair, however, and are unable to predict hairstyles from images.
\citet{sklyarova2023neural} proposed a strand-level hair reconstruction by optimization with volumetric constraints and priors learned from synthetic data and \citet{shen2023CT2Hair} leveraged computed tomography to create density volumes of the hair regions, though both methods require significantly more input data than a single image. 
\citet{Zheng_2023_CVPR} proposed a framework that maps an image to intermediate 2D strand and depth maps to inform 3D style reconstruction based on synthetic priors, though with no consideration of frizzy or coily hair. 

While conceptually attractive, direct parametric representation of hair strands needs significant additional work before it can underpin an inclusive hair prediction system. 
Styles considered are almost exclusively long, straight or wavy with limited mention of coily and short hair, and most parametric formulations are simply incompatible with frizzy or coily hair. 
Even recent methods designed to tackle this issue specifically~\cite{zhou2023groomgen} are unable to model very tightly coiled hair such as dreadlocks.
As a result, we instead focus on a hairstyle classification approach as we \emph{have} to be able to represent common styles including dreadlocks, buns and braids which current parametric models cannot.
We recognize that any form of classification of human attributes carries the risk of introducing biases to methodology and results \cite{lasisi2021constraints}. 
However, we show that with attention to asset definition and diversification, our approach can effectively handle a wide range of hairstyles.


\section{Method}
We train a DNN to predict hairstyle classes given a single input image of a person.
For training data we use synthetically rendered images with hairstyles sampled from a synthetic hairstyle library, the model predicts the hairstyle class present in the image, as well as associated taxonomy attributes as an auxiliary task.
For evaluation we use real image data to get an indication of the generalization of the model, which is not possible when evaluating on synthetic data.
This section describes the key technical components of our system: the hairstyle taxonomy, datasets, and prediction methodology.

\subsection{Hairstyle Taxonomy}

An accurate and fair system for predicting hairstyles requires a careful and systematic approach to data acquisition, labeling, model validation, and description of failure conditions. 
This requires a hairstyle characterization system with the following properties, which we aim to satisfy with our proposed taxonomy: 
\begin{itemize}
\item \emph{Completeness}. All human hairstyles are captured to some degree of accuracy.
\item \emph{Fairness}. Inaccuracy in the description of an actual hairstyle using the taxonomy is similar across all demographics.
\item \emph{Granularity}. The categories are granular enough to distinguish between the majority of hairstyles. 
\item \emph{Simplicity of use}. Diverse users (researchers, groom artists, users of avatars  etc.) can all understand the taxonomy. 
\item \emph{Consistency}. A single hairstyle is described by a unique combination of attributes. 
\item \emph{Objectivity}. The language refers to physical attributes rather than cultural references and is unambiguous.
\item \emph{Extensibility}. Allowing diverse communities to contribute to continuous taxonomy improvements.
\end{itemize}

\paragraph{Summary}
Hair type is reasonably well described in literature \cite{andrewalker,shapevariability,Curlcentric}.
We adopt the four types of \citet{andrewalker}, though use named categories rather than numbered ones \cite{counteringracial}.
Type, however, is only a very small part of the information required to determine a hair\emph{style}.
The remainder of our taxonomy is devised to cover the variation in \emph{styling} of human hair.
An overview of existing taxonomies and datasets is provided in the supplementary material.

Our hairstyle taxonomy consists of 18 attributes. 
There are ten global attributes which are based on the whole hairstyle, for example the shape of the hairline or surface appearance of the hair.
The scalp is divided into eight regions and each region is annotated with eight local attributes, such as length and strand styling.
So in total each hairstyle has $A=74$ taxonomic labels.
A graphical representation of the full taxonomy is shown in \autoref{fig:hair-taxonomy}, and a detailed textual description is given in the supplementary material.
While we hope that the taxonomy presented is sufficiently fair, objective and complete, we recognize that it is likely impossible for it to be \emph{truly} complete.
We therefore encourage future work to extend the taxonomy as required and publish any modifications.

\begin{figure*}[p]
    \centering
    \includegraphics[angle=270]{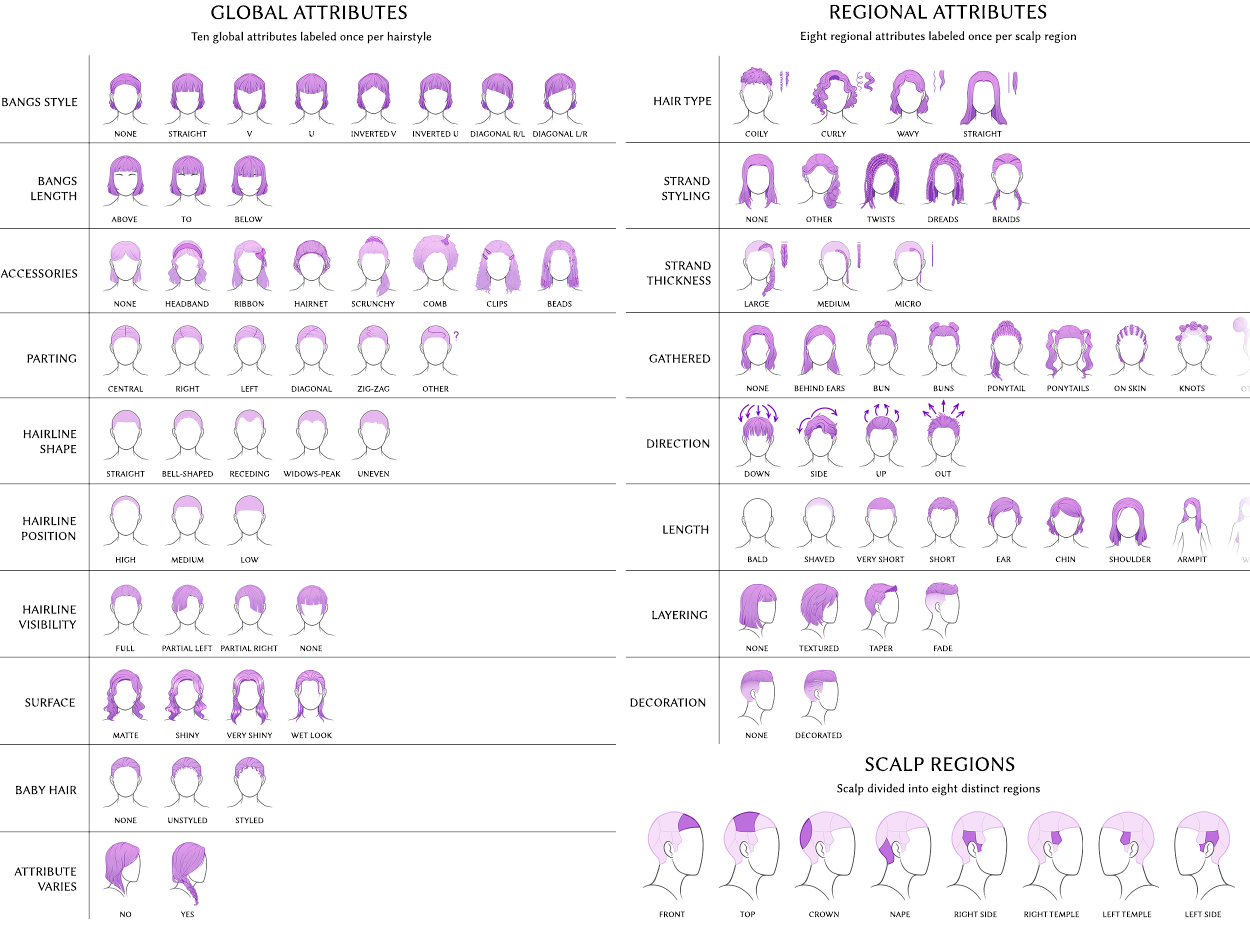}
    \caption{Graphical overview of our proposed hair taxonomy consisting of global and regional attributes. Note that some values are not visualized for the `Gathered' and `Length' attributes.}
    \label{fig:hair-taxonomy}
\end{figure*}

\paragraph{Design Process}
The taxonomy was designed in collaboration with domain experts including anthropologists to ensure objectivity in the terminology and categorization used. 
Scalp regions were adapted from those used by hair stylists and transplant specialists \cite{lee2007new, pascoe2023anatomy, blume2008hairgrowth}.
The taxonomy was iteratively refined through a series of pilot studies involving an ethnically diverse set of around 30 labellers from varying technical backgrounds (scientists, artists and annotators). 
To ensure ease of use and logical consistency, refinements for each iteration of taxonomy itself and the instructions were informed by evaluations of consistency and confidence of annotations in the earlier stages, and automatic rule-based analysis in the later stages.
This iterative refinement enabled an equitable taxonomy that achieves a good balance between simplicity and granularity.

\subsection{Datasets}
\label{sec:datasets}

To train our method we use only synthetic data.
This provides benefits such as elimination of privacy concerns commonly associated with human-centered research~\cite{bae2023digiface} and noise-free ground-truth annotation~\cite{wood2021fake}.
For our task specifically, it also significantly reduces annotation cost; annotation with the hairstyle taxonomy is relatively complex, but as we only need to annotate each \emph{asset} and not each \emph{image} the annotation cost scales with the asset library and not the dataset.
We are also able to explicitly control the diversity of the dataset by adjusting the sampling strategy for hairstyles in the generated scenes using taxonomic labels.
This allows us to reduce or eliminate bias through iterative refinement which is prohibitively expensive for most real-world data collections.

For evaluation we use a real-image dataset, this is critical to ensure generalization of our model to real-world data and highlights any issues of domain gap.
Using real data for evaluation only means collection and annotation costs are significantly lower as a much smaller volume of data is required.
We use taxonomic annotations for this real test set, which enables us to label real-image data with rich \emph{and accurate} ground-truth. 
This would simply not be possible if labeling directly using hairstyles from our synthetic library, as would be typical for a classification task.
For instance, for a given image, there may not be an appropriate style in our synthetic hairstyle library, resulting in approximate labels with a high degree of noise, introducing subjectivity and potential bias to the ground-truth annotations.
Taxonomic labels also mean we can easily update our synthetic hairstyle library, and so the classification results, without invalidating the ground-truth annotations on real data.
This lets us easily assess improvements to the library, or even evaluate totally different methods using the same metrics.

Both datasets are available to download from the project page: {\color{niceblue}\url{https://aka.ms/hairmony}}.

\begin{figure}
    \centering
    \includegraphics[width=\linewidth]{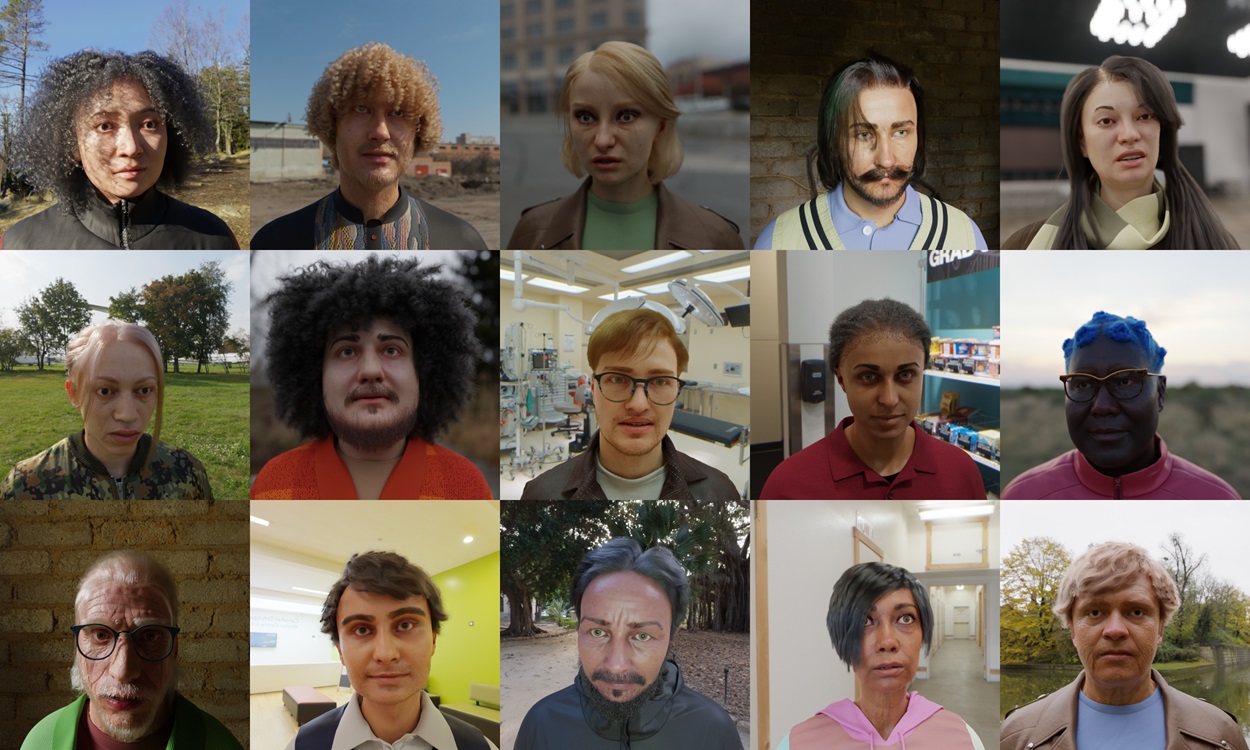}
    \caption{Example images from our synthetic training data showing a variety of hairstyles combined with random facial appearance, pose and environment.}
    \label{fig:synth-data}
\end{figure}

\paragraph{Synthetic Image Training Set} 
We annotate the synthetic hairstyle library of \citet{wood2021fake} using our taxonomy, such that each style, $s$, has associated taxonomy attributes, $\left[a_1^{s}, a_2^{s}, \ldots, a_{A}^{s}\right]$. 
We expand the library to address any common taxonomic combinations that were missing or under-represented.
In total, the expanded library contains 480 distinct hairstyles, represented as 3D strands.
We use the rendering pipeline of \citet{wood2021fake} to render a dataset containing images of the face with our expanded hair library.
The dataset contains 100,000 $512\times512$ pixel images rendered under different lighting conditions with a different pose and facial expression and randomly sampled camera positions.
Each synthetic image therefore has a hairstyle class annotation and the taxonomic annotations inherited from that hairstyle.
Hairstyles are sampled non-uniformly to promote fairness across various taxonomic attributes, as detailed below.
Some example training images are shown in \autoref{fig:synth-data}.

\paragraph{Real Image Evaluation Set} 

We annotate a subset of the FairFace dataset~\cite{karkkainen2019fairface} balanced across demographics with our taxonomy. 
The subset was selected to include images with a frontal view of a single person and free of occlusions while maintaining a balance of fairness attributes, see the supplementary material for details. 
In total our evaluation set contains 1805 samples.


\subsection{Hairstyle Sampling}

Following addition of hairstyles to the library to fill any obvious gaps in attribute combinations, we still find that the library is imbalanced in terms of some taxonomic attributes.
For a typical classification problem it would be standard to uniformly sample the classes, i.e., the styles from our hairstyle library, but this would propagate any imbalances in the hairstyle library to the training dataset and result in potential bias in the resulting models.

We instead choose to balance our synthetic dataset by weighting the sampling of hairstyles in a non-uniform way.
In so doing we promote fairness of the dataset along important, semantically meaningful attributes of hair, such as length and hair type.
Our balanced hairstyle attribute distribution is: 50\% with fringes; 75\% with gathering (in the back, in cornrows, behind the ear for long hairstyle); Short\slash Medium\slash Long hair: 40\%\slash30\%\slash30\%; Straight\slash Wavy\slash Curly\slash Coily: 50\%\slash15\%\slash15\%\slash20\% \cite{daniels2023different}.
The influence of the training set on different aspects of the trained model will be covered in \autoref{sec:results}.

\subsection{Network Architecture}

\begin{figure}
    \centering
    \includegraphics[width=\linewidth]{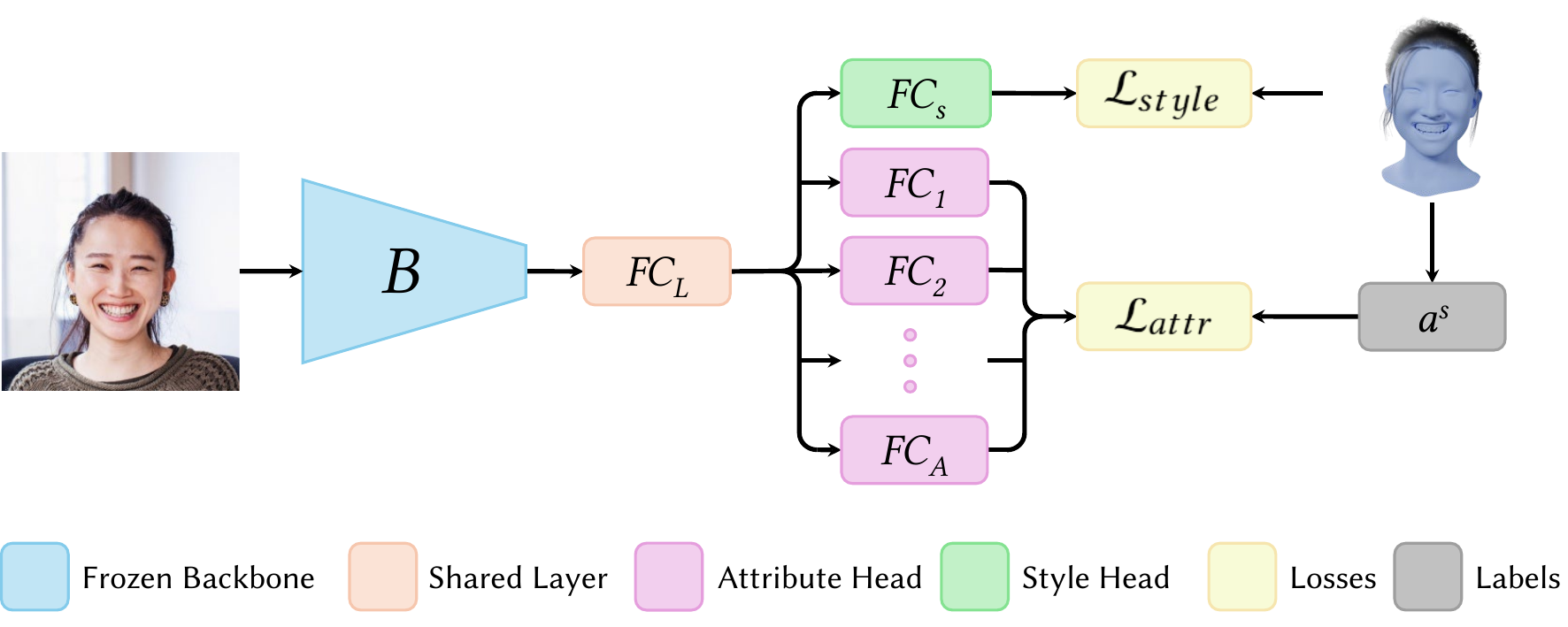}
    \caption{Training and operating scheme of the proposed model. Given an input image, frozen backbone $B$ produces intermediate features that are later used to produce intermediate representation via shared layer $FC_{L}$. Given the output of $FC_{L}$: \textit{\textbf{(i)}} hairstyle prediction head $FC_{s}$ predicts the final hairstyle and \textit{\textbf{(ii)}} hairstyle attribute prediction heads $FC_{1}, \ldots, FC_{A}$ predict taxonomic hairstyle attributes.}
    \label{fig:method}
\end{figure}

We use a convolutional neural network (CNN) trained entirely on synthetic data.
The network is comprised of a frozen backbone that has been pre-trained through self-supervision on a large corpus of real images to provide general-purpose visual features, DINOv2 followed by a number of fully-connected layers optimized during training, visualized in \autoref{fig:method}.

We find using a frozen, pre-trained backbone to be highly beneficial for the task of hairstyle prediction.
Given the relatively small library of synthetic hairstyles, and limited visual quality of synthetic images it is very easy for neural networks to over-fit when trained exclusively on our synthetic data. 
High frequency details are also important to determine details of hair type and style, but the domain gap between real and synthetic images may hamper the ability of CNNs to learn these details when trained only on synthetic data.
By using a model pre-trained on real images we minimize the ability of the network to over-fit, as we only optimize small fully connected layers, and ensure that we are extracting features that generalize to real images.

The primary task of our network is hairstyle prediction, formulated as classification task for hairstyles from our synthetic library.
We also include an auxiliary task of hairstyle attribute prediction, outputting the taxonomic annotations associated with the hairstyle.
Attributes are predicted by dedicated fully-connected heads which take shared features from a common fully-connected layer as input.
This architecture ensures that features from this intermediate layer are informed by attributes that we know are important (hair type, length, etc.) as determined by the taxonomy, rather than features indirectly inferred from a complex classification task.
The aim of this approach is to prevent `bad' errors; that is to ensure that even if our method does not predict the perfect style, it at least predicts a style with matching attributes.
A classification-only approach has no concept of this semantic similarity.
\label{sec:dnn_architecture}

Given an input facial image $x$ we first pass it through the DINOv2~\cite{oquab2023dinov2} backbone $B$ and obtain the latent feature $f=B(x)$, defined as:
\begin{align*}
    f&\equiv\Big[(E^{N}_{cls},E^{N}_{ptc}),(E^{N-1}_{cls},E^{N-1}_{ptc}),(E^{N-2}_{cls},E^{N-2}_{ptc}),(E^{N-3}_{cls},E^{N-3}_{ptc})\Big]
\end{align*}
where 
$E$ denotes output tokens of DINOv2, superscripts define the DINOv2 layer ($N$ indexes the last layer), subscript $cls$ denotes DINOv2 class tokens, subscript $ptc$ denotes average pooled patch tokens. Both class and patch tokens $(E_{cls},E_{ptc})$ are used, to ensure that our classifier  has access to both high-level semantic and low-level appearance information.
The features $f$ from backbone $B$ are further processed by one fully-connected layer, $FC_{L}$ producing the intermediate representation $l=FC_{L}(f)$. 
Latent feature $l$ is shared between final hairstyle prediction head, $\hat{s}=FC_{s}(l)$, and taxonomic attribute prediction heads, $\hat{a}_{1}^{s}=FC_{1}(l), \ldots, \hat{a}_{A}^{s}=FC_{A}(l)$. 
Hairstyle prediction head $FC_{s}$ and taxonomic attribute prediction heads, $FC_{1}, \ldots, FC_{A}$, consist of one fully-connected layer networks. 
Taxonomic attribute prediction during training attempts to maximize the mutual information between predicted hairstyle $\hat{s}$ and corresponding taxonomic attributes $[\hat{a}_{1}^{s}, \ldots, \hat{a}_{A}^{s}]$.



\subsection{Training Losses}
\label{sec:losses}
Given an input selfie image, during training we optimize the model to predict the distribution over our library of hairstyles and associated taxonomic labels. 
Given intermediate feature $l = FC_{L}(B(x))$ for input image $x$ we evaluate two losses.
\paragraph{Hairstyle Loss} 
We supervise the hairstyle prediction head $FC_{s}$ by minimizing the cross-entropy loss between predicted distribution of assets and one-hot encoding of the ground-truth hairstyle, $s$:
\begin{align}
\mathcal{L}_{style}(l,s) = \mathcal{L}_{ce}(FC_{s}(l)),s)
\label{eq:style_cross_entropy}
\end{align}
\paragraph{Attribute Loss} 
As mentioned in \autoref{sec:dnn_architecture}, in addition to predicting synthetic hairstyles from our library, we also predict semantic hairstyle attributes, $\left[\hat{a}_1^{s}, \ldots, \hat{a}_{A}^{s}\right]$, as defined by our taxonomy. 
Hence, we define an additional attribute related loss term $\mathcal{L}_{attr}$ as follows:
\begin{align} 
\mathcal{L}_{attr}(l,s)=\sum_{t=1}^{A} \mathcal{L}_{ce}(FC_{t}(l),a_{t}^{s})
\label{eq:attribute_cross_entropy}
\end{align}

During training we optimize the parameters of all fully connected layers to minimize the sum of $\mathcal{L}_{style}$ and $\mathcal{L}_{attr}$, while DINOv2 backbone, $B$, remains frozen.

\section{Experiments}
\begin{figure*}[p]
    \centering
    \includegraphics[width=\linewidth]{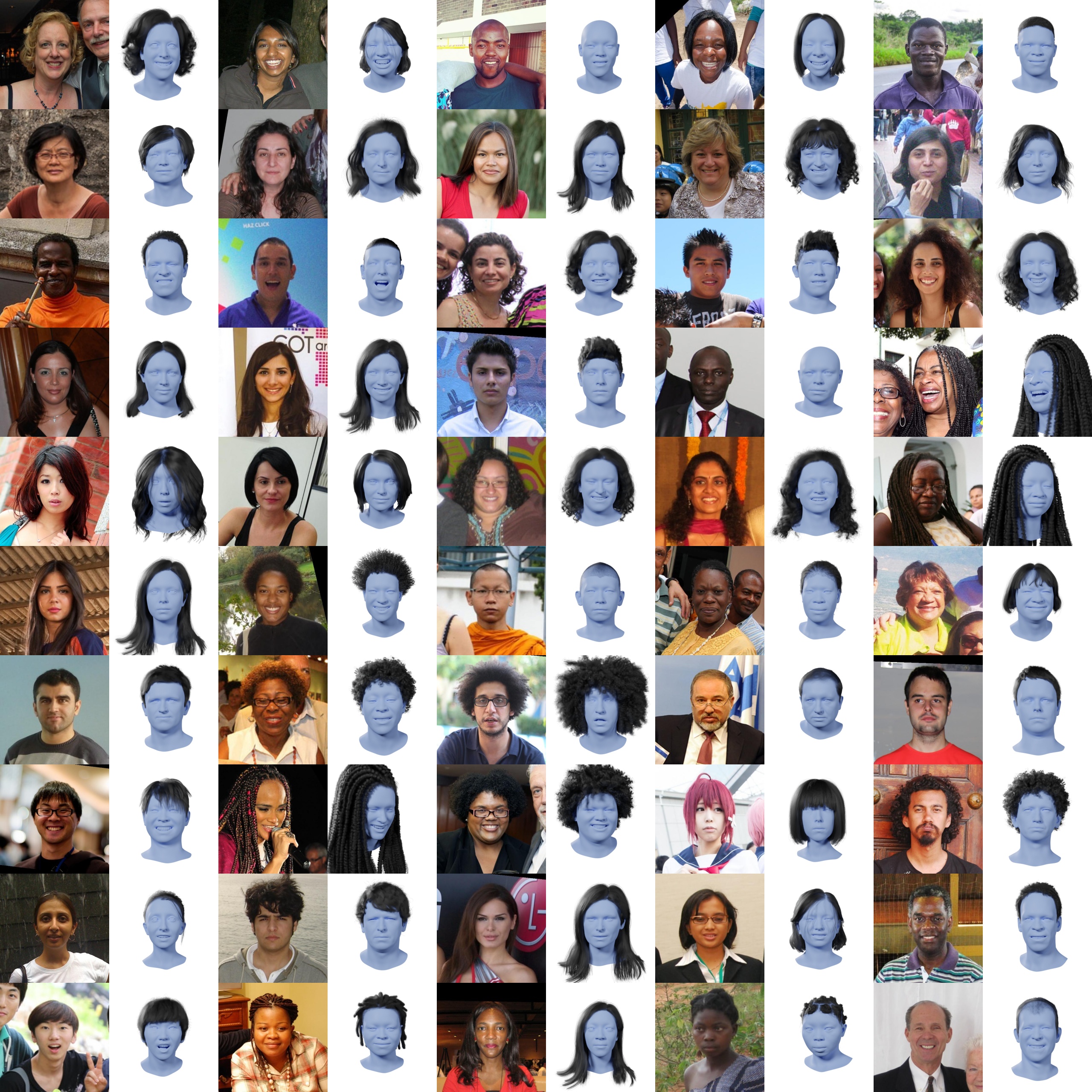}\\
    ~\\
    \includegraphics[width=\linewidth]{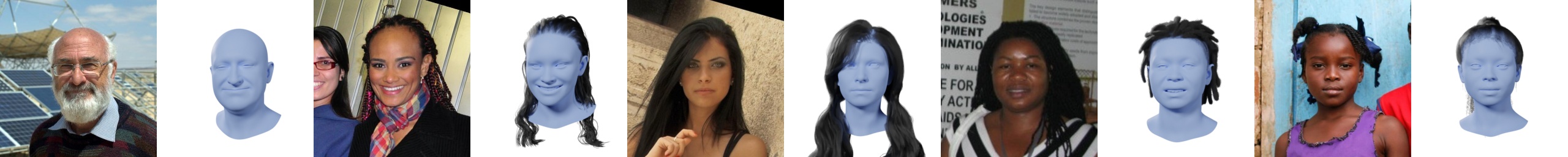}
    \caption{Qualitative results for our method on the FairFace~\cite{karkkainen2019fairface} evaluation subset. 
    Bottom row shows failure cases, specifically: missed hair, incorrect hair type/strand styling, incorrect gathering, incorrect length, hairstyle not in library. 
    \fairfacecopy}
    \label{fig:qual-results}
\end{figure*}

To evaluate our model we conduct a number of experiments assessing both accuracy \emph{and} fairness.
\begin{figure*}
    \centering
    \footnotesize
    \begin{tabularx}{.49\linewidth}{YYYYYY}
        Input & HairStep & HairStep & HairStep & Ours & Ours\\
        Image & (strands) & (front) & (side) & (front) & (side)
    \end{tabularx}\hfill\begin{tabularx}{.49\linewidth}{YYYYYY}
        Input & HairStep & HairStep & HairStep & Ours & Ours\\
        Image & (strands) & (front) & (side) & (front) & (side)
    \end{tabularx}
    \includegraphics[width=0.49\linewidth]{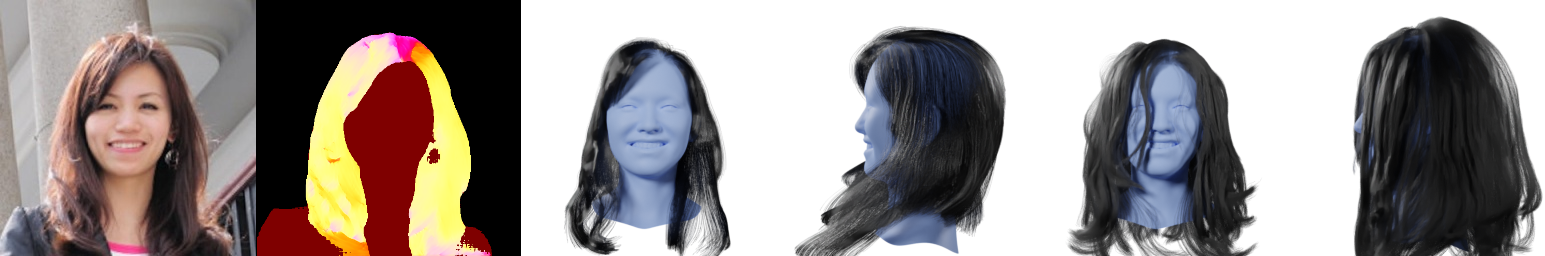}\hfill\includegraphics[width=0.49\linewidth]{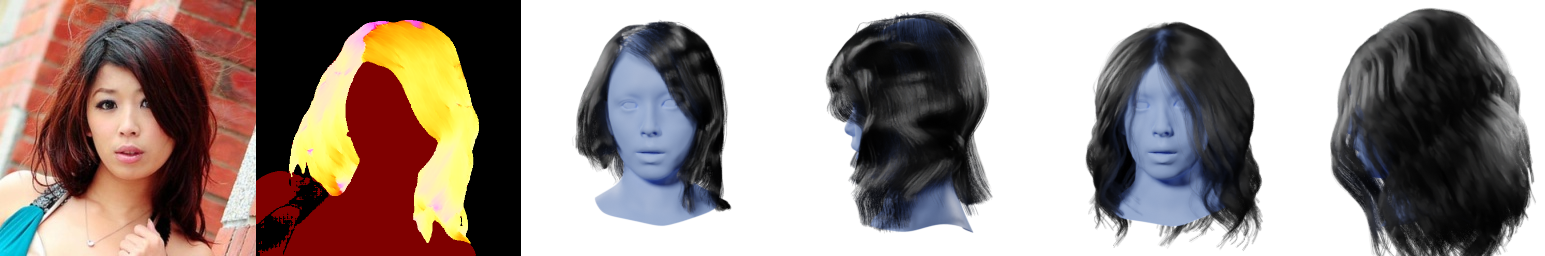}\\
    \includegraphics[width=0.49\linewidth]{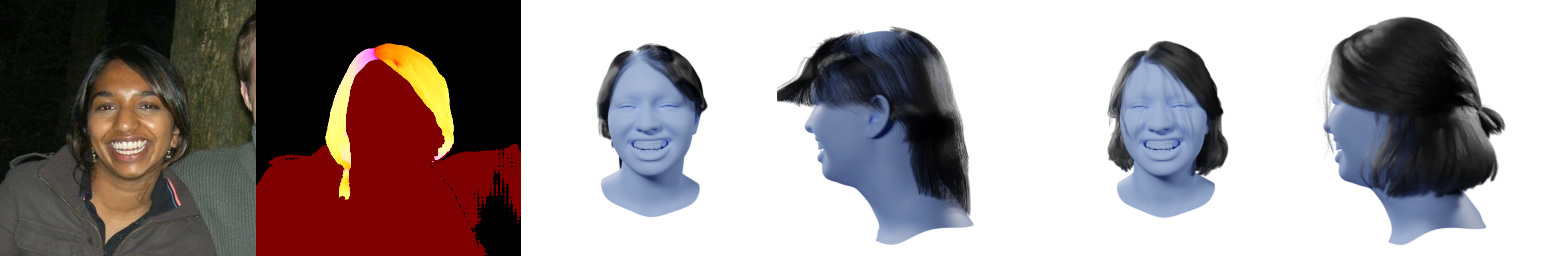}\hfill\includegraphics[width=0.49\linewidth]{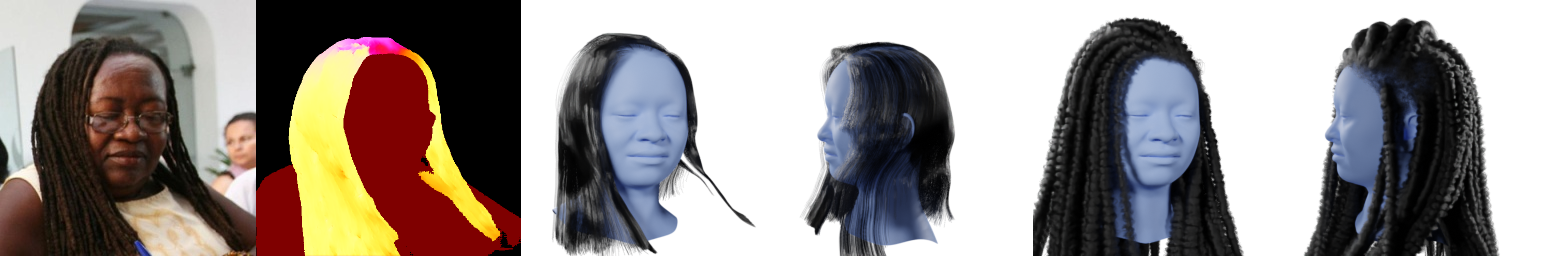}\\
    \includegraphics[width=0.49\linewidth]{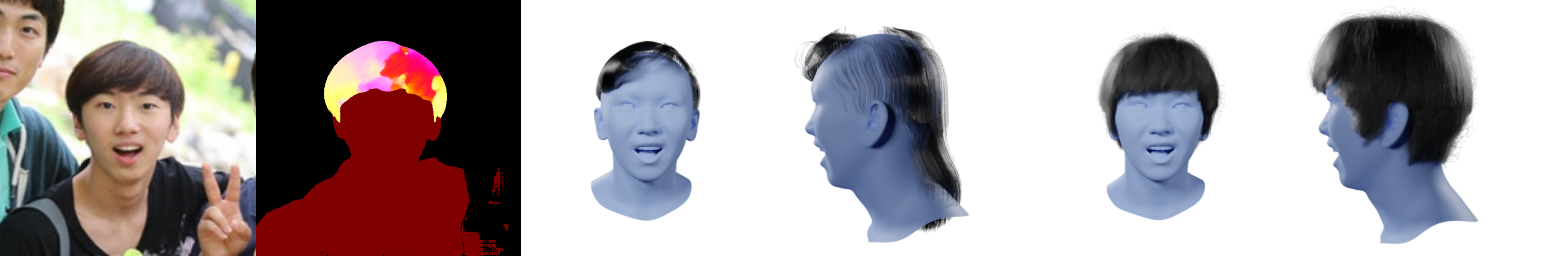}\hfill\includegraphics[width=0.49\linewidth]{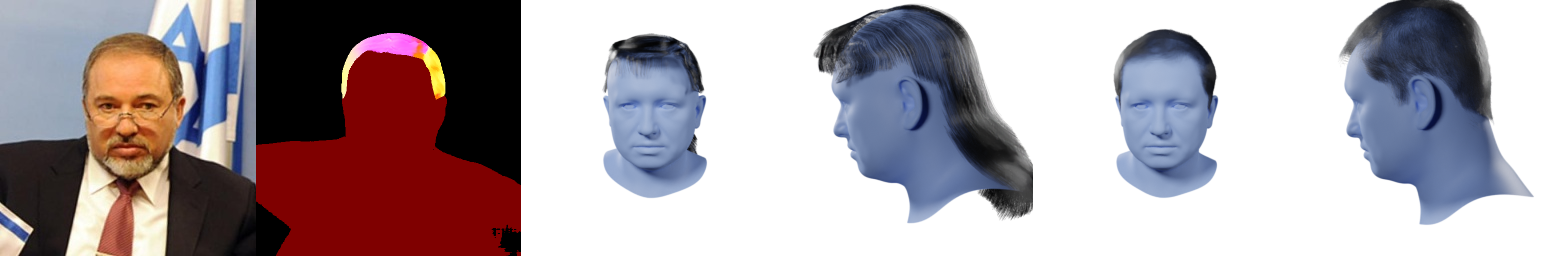}\\
    \includegraphics[width=0.49\linewidth]{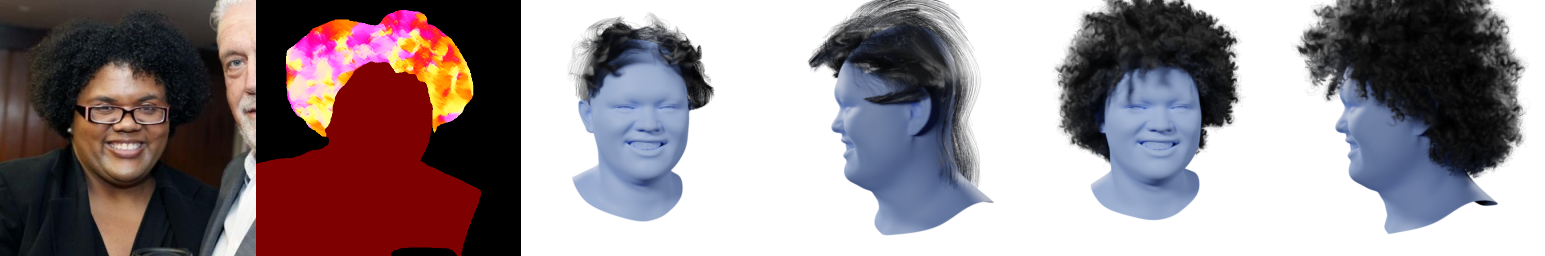}\hfill\includegraphics[width=0.49\linewidth]{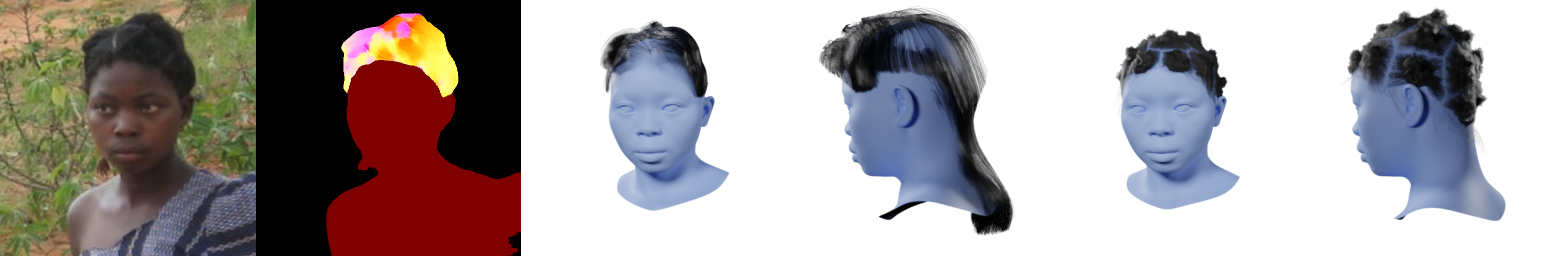}\\
    \caption{Comparison of our method with recent parametric hair prediction approach HairStep~\cite{Zheng_2023_CVPR}.
    HairStep performs well for long straight hair (top row), but has a strong bias towards this hair style and type.
    This results in poor performance for short styles and coily or curly hair types, even if results appear to be of reasonable quality when viewed from the front. While our results provide less direct representation in some cases, they are significantly more robust across diverse hairstyles. 
    HairStep results are manually aligned to reconstructed face meshes. \fairfacecopy}
    \label{fig:sota_comp}
\end{figure*}
As we cannot compare to recent parametric approaches quantitatively without manual labeling of generated results with our taxonomy, we provide a qualitative comparison of our method with the state-of-the-art method for hairstyle reconstruction, shown in \autoref{fig:sota_comp}.
While our method does not enable direct strand-wise representation, it is far more robust for diverse input hairstyles.
Existing methods show a strong bias to straight, long hair while our method is able to provide appropriate hairstyle predictions for short, frizzy, coily and gathered styles, as well as long, straight hair.
Further qualitative results of our method are shown in \autoref{fig:teaser} and \autoref{fig:qual-results}, face shapes are reconstructed using method of \citet{wood20223d}.
The remainder of this section describes the experimental setup, evaluation metrics and quantitative results.

\subsection{Training Setup}
\label{sec:experimental_setup}
The DINOv2 backbone is frozen, shared layer $FC_{L}$, hairstyle prediction layer $FC_{s}$, and taxonomic attribute prediction layers $FC_{a}, a\in[1,\ldots,A]$ are jointly trained for 30 epochs on the synthetic training set containing 100K images. 
$FC_{L}$ has input dimension 8192 and output dimension 4096 with ReLU activations and dropout.
$FC_{s}$ has input and output dimensions of 4096 and 480 respectively. 
Layers $FC_{a}$ have input dimension 8192 and output dimensions matching the number of taxonomic attribute values for each attribute.
We use the AdamW optimizer \cite{loshchilov2018decoupled} with initial learning rate of $3\times10^{-4}$ decayed using a cosine annealing scheduler \cite{loshchilov2017sgdr}. 
Our models are trained on 4 NVidia A100 GPUs, with a batch size of 512. 
Executing a forward pass of our model for a single image on an NVIDIA A100 GPU requires 11 milliseconds.

\subsection{Evaluation Metrics}
We assess the quality of our trained model on a labeled subset of the FairFace dataset~\cite{karkkainen2019fairface} described in \autoref{sec:datasets}.
We compute both accuracy \emph{and} fairness for a number of taxonomy attributes. 
We focus on the following set of attributes: bald, bang styling, hair gathered, hair length, hair type, and strand styling. 
Bald is a subset of the hair length attribute. 
This set of metrics was chosen based on correlation with protected demographic groups, and consequent significance for fairness.
This is in-line with previous attribute and hairstyle classification work (\autoref{sec:past_work_class}) with addition of the hair gathered attribute, a common failure case in existing methods. 
Many attributes in our taxonomy include regional annotations so reporting individually in a concise manner is challenging, we therefore collate labels from all scalp regions.
These taxonomy-based accuracy metrics serve as a \emph{perceptual} measure of hairstyle similarity which are agnostic to the exact hair classes being predicted, providing a good indication of acceptability of the predicted style to a user. 
To calculate fairness for three most common demographics categories (gender, age, ancestry), we first calculate attribute accuracy across each group within a category (e.g. female/male for gender). Then we calculate attribute fairness as mean accuracy across different demographics groups divided by the maximum accuracy of any individual group and we represent it as percentage. This provides metric that converges to 100\% as disparity between groups becomes smaller.
\label{sec:metrics}

\subsection{Experimental Results}
\label{sec:results}

\paragraph{Baselines} 
\autoref{tab:ablations} reports classification accuracy for different hairstyle attributes and mean fairness over different models. 
Prior work performing hairstyle classification or attribute prediction often follows conventional machine learning methods, varying in specifics based on the data formats used \cite{yin2017learning, hu2017avatar, svanera2016figaro,chen2021celebhair,kim2021k}.
To approximate these we choose a ResNet34~\cite{he2016deep} baseline trained using $\mathcal{L}_\text{style}$, i.e., conventional cross-entropy loss (Row 1).
In rows 2-4, we report results of our DINOv2-based models.
All models were trained to classify hairstyles into the 480 hairstyles from our catalog. 
An extended version of \autoref{tab:ablations} with additional results for ResNet50 and ResNet101 models can be found in supplementary material.

\paragraph{Effect of the DINOv2 backbone on accuracy}
To assess over-fitting we use an additional synthetic test dataset of 10000 images.
The ResNet34 baseline model achieves $97.7\%$ training accuracy and $93.7\%$ accuracy on the additional synthetic test set, suggesting that the model has over-fit to the synthetic training images.

For comparison, the DINOv2-based model achieves a training accuracy of $74.8\%$ and $78.0\%$ accuracy on the additional synthetic test set.
As shown in rows 1-2 of the \autoref{tab:ablations}, the DINOv2-based model significantly outperforms the baseline model on the real-image test set.
This demonstrates that by preventing over-fitting to the synthetic data, the model is better able to generalize to real images.

\newcommand{\TA}[1]{\cellcolor{green!25}\textbf{#1}}
\newcommand{\TB}[1]{\cellcolor{yellow!25}#1}
\newcommand{\TC}[1]{\cellcolor{orange!25}#1}
\newcommand{\TD}[1]{\cellcolor{red!25}#1}

\begin{table*}
    \centering
    \footnotesize
    \caption{
    Results of our ablation experiments. We report hairstyle attribute classification accuracy for \textbf{\textit{(i)}} Baseline ResNet34 \cite{He_2016_CVPR} architecture trained with $\mathcal{L}_{style}$ loss in row 1; \textbf{\textit{(ii)}} Model with frozen DINOv2 \cite{oquab2023dinov2} backbone trained with $\mathcal{L}_{style}$ in row 2; \textbf{\textit{(iii)}} Model with frozen DINOv2 backbone trained with $\mathcal{L}_{style}$, and $\mathcal{L}_{attr}$ losses in row 3; \textbf{\textit{(iv)}} Our final model with frozen DINOv2 backbone trained with $\mathcal{L}_{style}$, and $\mathcal{L}_{attr}$ losses on attribute balanced training set.
    }
    \begin{tabular}{lcccccccc}
        \toprule
         \textbf{Method} & \textbf{Bald} & \textbf{Bang Styling}  & \textbf{Gathered} & \textbf{Length} & \textbf{Hair Type} & \textbf{Strands} & \textbf{Mean Accuracy} & \textbf{Mean Fairness}\\
        \midrule
        Resnet34 + $\mathcal{L}_{style}$ & \TD{95.6\%} & \TB{84.3\%} & \TD{77.1\%} & \TD{63.7\%} & \TD{81.8\%} & \TD{95.2\%} & \TD{83.0\%} & \TA{92.9\%}\\
        Ours + $\mathcal{L}_{style}$ & \TC{96.8\%} & \TD{83.0\%} & \TB{86.2\%} & \TC{69.6\%} & \TA{90.9\%} & \TB{98.5\%} & \TC{87.5\%} & \TC{91.9}\%\\
        Ours + $\mathcal{L}_{style}$ + $\mathcal{L}_{attr}$ & \TB{96.9\%} & \TA{84.4\%} & \TA{87.1\%} & \TA{70.6\%} & \TC{90.1\%} & \TA{98.7\%} & \TA{88.0\%} & \TD{91.7\%} \\
        Ours + $\mathcal{L}_{style}$ + $\mathcal{L}_{attr}$ Balanced & \TA{98.0\%} & \TC{83.8\%} & \TC{85.4\%} & \TB{69.8\%} & \TB{90.2\%} & \TB{98.5\%} & \TB{87.6\%} & \TB{92.5\%}\\
        \bottomrule
    \end{tabular}
    \label{tab:ablations}
\end{table*}
\begin{figure*}
    \centering
    \includegraphics[width=\linewidth]{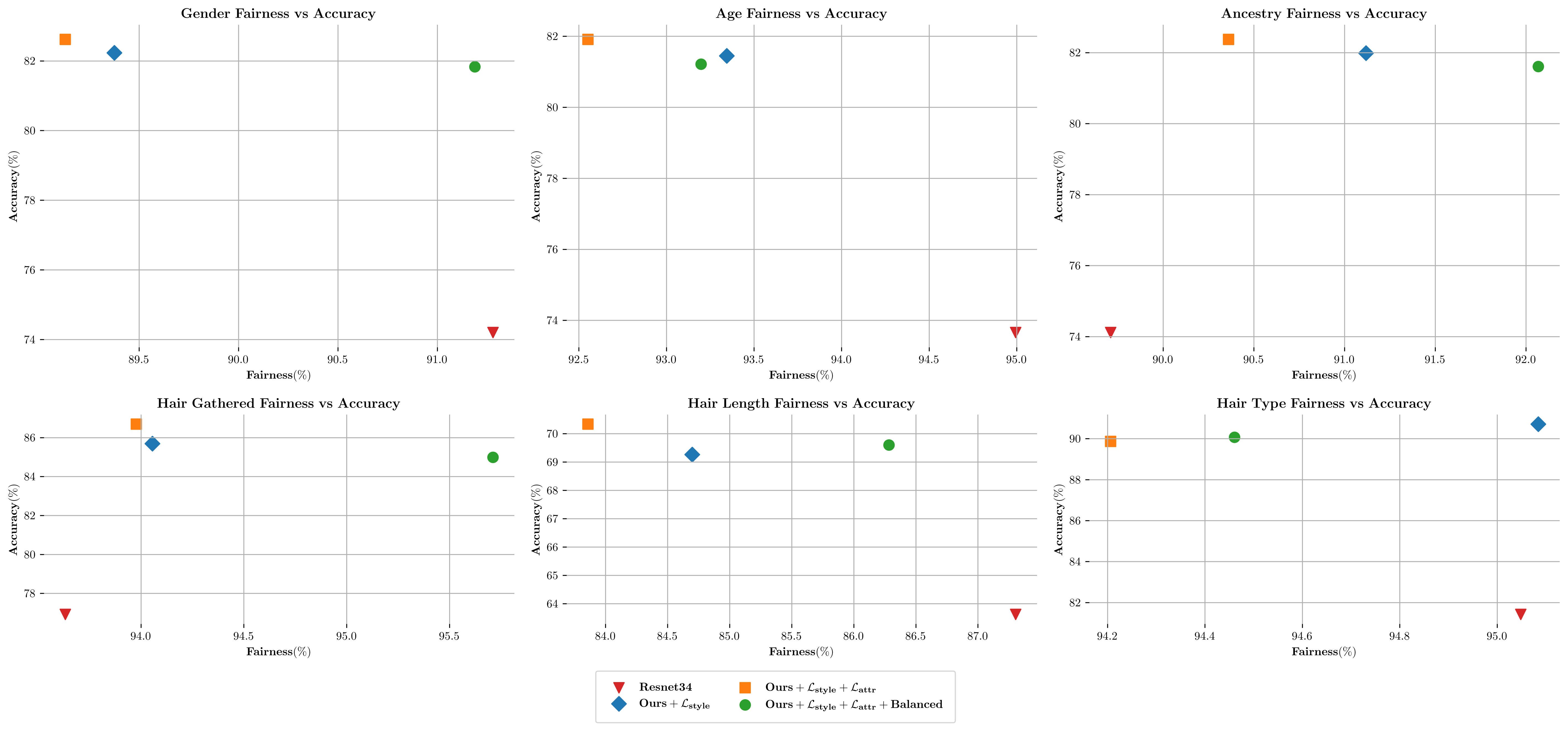}
    \caption{Visualization of algorithmic fairness versus hairstyle attribute prediction accuracy for our our final model and baseline models. Models achieving high accuracy and fairness simultaneously (our target) should appear in the top right corner of the plot. The first row contains results for gender, age, and ancestry groups averaged across different attributes. The second row contains results for hair gathered, hair length, and hair type metrics averaged over different demographic groups (gender, age, and ancestry). In most cases our final model, shown in green, achieves the best combination of fairness and accuracy.}
    \label{fig:fairness_ablations}
\end{figure*}

\paragraph{Effect of auxiliary hairstyle attribute prediction} 
We investigate the influence of auxiliary hairstyle prediction loss term, $\mathcal{L}_{attr}$, defined in \autoref{eq:attribute_cross_entropy}. 
In the third row of \autoref{tab:ablations} we report the results of the model trained to minimize the $\mathcal{L}_{style}$ and $\mathcal{L}_{attr}$ losses simultaneously. 
One can observe that hairstyle attribute prediction overall improves the accuracy for most metrics.
\paragraph{Controlling fairness through training data statistics} 
We investigate how the distribution of different hairstyle attributes in training data influences the fairness of the trained model. 
\autoref{tab:ablations} shows the impact on accuracy of training with the balanced dataset compared with a dataset using uniform sampling (bottom two rows).
Balancing the dataset results in a small degradation in accuracy, however, as shown in \autoref{fig:fairness_ablations}, results in a significant improvement in fairness across the majority of groups and attributes, while retaining similar accuracy. In particular we observe a significant fairness improvement for the hair gathered attribute across gender from $85.3\%$ to $91.7\%$, a particularly challenging attribute due to visual ambiguity between gathered and short hairstyles in frontal images. A table with fairness results for different models can be found in the supplementary material.

\section{Limitations and Future Work}
Example failure cases are shown in the bottom row of \autoref{fig:qual-results}. 
We observe that in very dark or bright illumination conditions our model confuses coily hair with other hair types, which reduces fairness and accuracy for certain ancestry groups dominated by coily hair.
Sometimes our model confuses long, gathered hair with short and medium-length, non-gathered hair due to the absence of visual cues in the image. 
These issues can be partially addressed via further resampling training data and incorporating training samples that better cover the aforementioned scenarios. 
Attributes such as hair gathering, however, are challenging due to the lack of visual information for the back of the head.
Multiple input images could provide a solution to this problem.
In the long term, parametric modeling of hairstyles is likely necessary and should remain an active area of research.
To enable inclusive products in the mean-time, classification approaches are likely to lead to better outcomes for a broader range of users.



Our taxonomy was designed with attention to equitable language and a large gamut of hairstyles. 
However, given the constantly evolving nature of language and hairstyles, as well as the potential for different applications of the taxonomy, we recognize that refinements will be necessary. 
For instance, beauty applications may require more detailed descriptions of curliness and physical attributes of hair. 
Moreover, it is crucial to acknowledge that any classification of human attributes carries the risk of unethical applications and biases \cite{lasisi2021constraints, genderasvariable}. 
To foster diverse input and mitigate potential harm, we share the taxonomy as an open-source asset. 




\section{Conclusion}

In this work we have introduced a novel hairstyle taxonomy developed in collaboration with diverse domain experts. 
The taxonomy enables objective and granular description of a wide variety of hairstyles and provides scope for extension to ensure fairness and, as far as is possible, completeness.
We use this taxonomy to generate a balanced synthetic dataset with hairstyle annotations which we use to train a hairstyle classification model.
To improve generalization of this model to real image data we use a frozen CNN backbone pre-trained on real image data, and taxonomy attribute prediction as an auxiliary task with shared feature extraction layers.
We evaluate our model on a real-image test set using metrics derived from semantically meaningful taxonomy attributes to ensure both high accuracy \emph{and} fairness of our model.
We show that our method outperforms baseline methods and demonstrate it to be significantly fairer and more robust when compared qualitatively with the state-of-the-art parametric hairstyle reconstruction approach.

\appendix

\section{Existing Taxonomies and Datasets}
\autoref{tab:taxonomy_comp} provides an overview of existing taxonomy for hair type, note that these do not cover hair \emph{styling}. 
\autoref{tab:dataset_comp} gives a summary of existing datasets for hairstyle prediction and their annotation schemes.
These are often limited in terms of diversity of style, and richness of annotation.

\begin{table}[b!]
\centering
\caption{Popular hair typing systems.}
\scriptsize
\begin{tabularx}{\linewidth}{@{}ll@{}}
\toprule
\textbf{System name} & \textbf{Attributes} \\
\midrule
\citet{andrewalker} & \textbf{Curl Pattern}: \textsc{[1(a-c)-straight, 2(a-c)-wavy, 3(a-c)-curly, 4(a-c)-kinky]} \\
\midrule
\multirow{4}{*}{\parbox{1.75cm}{\citet{shapevariability}}} &  \textbf{Curve Diameter}: \textsc{[cm]} \\
& \textbf{Curl Index}: \textsc{[1]} \\
& \textbf{Number of waves}: \textsc{[1]} \\
& \textbf{Number of twists}: \textsc{[1]} \\
\midrule
\multirow{3}{*}{LOIS \cite{Curlcentric}} & \textbf{Curl Pattern}: \textsc{[L-bend, O-curl, I-straight, S-wave]} \\
& \textbf{Strand size}: \textsc{[thin, medium, thick]}\\
& \textbf{Texture}: \textsc{[thready, wiry, cottony, spongy, silky}] \\
\midrule
\multirow{3}{*}{FIA \cite{Curlcentric}} & \textbf{Follicle (Curl Pattern)}: \textsc{[1-straight, 2–wavy,3-curly, 4-really curly]}\\
& \textbf{Individual Strand Thickness}: \textsc{[fine, medium, coarse]}\\ &\textbf{All-Over Density}: \textsc{[thin, normal, thick]} \\
\bottomrule
\end{tabularx}
\label{tab:taxonomy_comp}
\end{table}
\begin{table}[b!]
\centering
\caption{Hairstyle datasets: Name, Number of hairstyles (if any), Discrete parameters (if any) excluding N/A, Other relevant annotations, Number of images if applicable and number of subjects in the dataset, if provided. }
\tiny
\resizebox{\linewidth}{!}{
\begin{tabularx}{\linewidth}{@{}lllll@{}}
\toprule
\textbf{Dataset} & \textbf{\# Styles} & \textbf{Discrete parameters} & \textbf{Other relevant} &  \textbf{\# Images}\\
\midrule
\multirow{3}{*}{\textbf{FACET} \citet{Gustafson_2023_ICCV}} & \multirow{3}{*}{-} & 6 $\times$ binary shape attr (2) & \multirow{3}{*}{Hair masks} & \multirow{3}{*}{31,702} \\
& & 6 $\times$ binary color attr (2) & & \\
& & 6 $\times$ accessory attr (6)\\
\midrule
\multirow{4}{*}{\textbf{K-Hairstyle} \citet{kim2021k}} & \multirow{4}{*}{31} & basestyle (31), length (5), & \multirow{4}{*}{-} & \multirow{4}{*}{500,000}  \\
& & basestyle\_type (2), curl (9), & & \\
& & bang (6), loss (4), side (2),\\
& & color (9), exceptional (7)\\
\midrule
\textbf{Figaro1k} \citet{svanera2016figaro} & - & \textsc{type (7)}  & - & 1,050 \\
\midrule
\multirow{3}{*}{\textbf{Beauty e-xpert}, \citet{Liu2013beautyexpert}} & \multirow{3}{*}{-} & length (3), shape (3), & \multirow{3}{*}{-} & \multirow{3}{*}{1,505} \\
& & volume (2), color (20), \\
& & bangs (4)\\
\midrule
\textbf{Hairstyle30K} \citet{yin2017learning} & 64 & - & - & 30,000\\
\midrule
\textbf{CelebHair} \citet{chen2021celebhair}   & 10  & - & - & 200,000 \\
\midrule
\textbf{CelebAMask-HQ} \citet{Lee2023celebamaskhq} & - & - & Hair masks & 30,000\\
\midrule
\multirow{2}{*}{\textbf{FairFace hair annotations (ours)}} & \multirow{2}{*}{-} & 10 global attributes & \multirow{2}{*}{-} & \multirow{2}{*}{1805}\\
& & 8 $\times$ 8 regional attributes \\
\midrule
\multirow{2}{*}{\textbf{Synthetic hair images (ours)}} & \multirow{2}{*}{480} & 10 global attributes & \multirow{2}{*}{Hairstyle name}  & \multirow{2}{*}{100,000} \\
& & 8 $\times$ 8 regional attributes &  \\
\bottomrule
\end{tabularx}
 }
\label{tab:dataset_comp}
\end{table}

\section{Dataset Details}
\autoref{tab:fairface_subset_stats} provides details of the statistics of the subset of the FairFace~\cite{karkkainen2019fairface} dataset we annotate with our taxonomy and use for evaluation.

\begin{table*}
    \centering
    \footnotesize
    \caption{Attribute counts on the labelled subset of FairFace \cite{karkkainen2019fairface}, real images used for evaluations.}
    \resizebox{\linewidth}{!}{
    \begin{tabular}{l|ccccccc|cc|ccccccc}
        \toprule
         & \multicolumn{7}{c}{\textbf{Age}} & \multicolumn{2}{c}{\textbf{Gender}} & \multicolumn{7}{c}{\textbf{Ancestry}}\\
         \midrule
         \textbf{Groups} & 10-19 & 20-29 & 30-39 & 40-49 & 50-59 & 60-69 & 70+ & Female & Male & Black & East Asian & Indian & Latino & Middle Eastern & Southeast Asian & White \\
         \textbf{Subjects} & 197 & 313 & 290 & 348 & 304 & 167 & 180 & 930 & 869 & 227 & 261 & 279 & 272 & 260 & 265 & 235 \\
         \bottomrule
    \end{tabular}
    }
    \label{tab:fairface_subset_stats}
\end{table*}
\begin{table}[!ht]
    \centering
    \footnotesize
    \caption{
    Results of our ablation experiments. We report hairstyle attribute classification accuracy for \textbf{\textit{(i)}} ResNet baseline \cite{He_2016_CVPR} architectures trained with $\mathcal{L}_{style}$ loss in rows 1-3; \textbf{\textit{(ii)}} Model with frozen  DINOv2 \cite{oquab2023dinov2} backbone trained with $\mathcal{L}_{style}$ in row 4; \textbf{\textit{(iii)}} Model with frozen  DINOv2 backbone trained with $\mathcal{L}_{style}$, and $\mathcal{L}_{attr}$ losses in row 5; \textbf{\textit{(iv)}} Our final model with frozen  DINOv2 backbone trained with $\mathcal{L}_{style}$, and $\mathcal{L}_{attr}$ losses on attribute balanced training set.
    }
    \resizebox{\linewidth}{!}{
    \begin{tabular}{lccccccc}
        \toprule
         \textbf{Method} & \textbf{Bald} & \textbf{Fringe}  & \textbf{Gathered} & \textbf{Length} & \textbf{Hair Type} & \textbf{Strands} & \textbf{Mean}\\
        \midrule
        Resnet50 + $\mathcal{L}_{style}$ & 95.6\% & 83.4\% & 75.0\% & 63.0\% & 82.5\% & 95.8\% & 82.6\% \\
        Resnet34 + $\mathcal{L}_{style}$ & 95.6\% & 84.3\% & 77.1\% & 63.7\% & 81.8\% & 95.2\% & 83.0\% \\
        Resnet101 + $\mathcal{L}_{style}$ & 95.7\% & 83.6\% & 75.5\% & 63.7\% & 83.1\% & 96.0\% & 82.9\% \\
        Ours + $\mathcal{L}_{style}$ & 96.8\% & 83.0\% & 86.2\% & 69.6\% & 90.9\% & 98.5\% & 87.5\% \\
        Ours + $\mathcal{L}_{style}$ + $\mathcal{L}_{attr}$ & 96.9\% & 84.4\% & 87.1\% & 70.6\% & 90.1\% & 98.7\% & 88.0\% \\
        Ours + $\mathcal{L}_{style}$ + $\mathcal{L}_{attr}$ Balanced & 98.0\% & 83.8\% & 85.4\% & 69.8\% & 90.2\% & 98.5\% & 87.6\% \\
        \bottomrule
    \end{tabular}
    }
    \label{tab:ablations}
\end{table}

\begin{table*}
    \tiny
    \centering
    \caption{Results of our ablation experiments. We report fairness for different attributes across different genders, age groups and ancestries for \textbf{\textit{(i)}} Baseline architectures trained with $\mathcal{L}_{style}$ loss in rows 1-3; \textbf{\textit{(ii)}} Model with frozen  DINOv2 backbone trained with $\mathcal{L}_{style}$ loss in row 4; \textbf{\textit{(iii)}}; Model with frozen  DINOv2 backbone trained with $\mathcal{L}_{style}$ and $\mathcal{L}_{attr}$ losses in row 5; \textbf{\textit{(iv)}}  Our final model with frozen  DINOv2 backbone trained with $\mathcal{L}_{style}$, and $\mathcal{L}_{attr}$ losses on attribute balanced training set.}
    \resizebox{\linewidth}{!}{
    \begin{tabular}{llccccccccccccccccc}
        \toprule
         & \multirow{2}{*}{\textbf{Method}} & \multicolumn{4}{c}{\textbf{Fringe fairness}} & \multicolumn{4}{c}{\textbf{Gathered  fairness}} & \multicolumn{4}{c}{\textbf{Length fairness}} & \multicolumn{4}{c}{\textbf{Hair Type fairness}} & \multirow{2}{*}{\textbf{Total Mean}} \\
         & & \textbf{Age} & \textbf{Gender} & \textbf{Ancestry} & \textbf{Mean} & \textbf{Age} & \textbf{Gender} & \textbf{Ancestry} & \textbf{Mean} & \textbf{Age} & \textbf{Gender} & \textbf{Ancestry} & \textbf{Mean} & \textbf{Age} & \textbf{Gender} & \textbf{Ancestry} & \textbf{Mean} & \\
        \midrule
 & Resnet50 + $\mathcal{L}_{style}$ & 96.88\% & 99.05\% & 87.16\% & 94\% & 94.14\% & 94.15\% & 93.6\% & 94\% & 87.26\% & 86.53\% & 89.25\% & 88\% & 95.7\% & 97.0\% & 94.1\% & 96\% & 92.9\% \\
 & Resnet34 + $\mathcal{L}_{style}$ & 98.85\% & 100.0\% & 88.5\% & 96\% & 94.98\% & 92.99\% & 92.93\% & 94\% & 93.13\% & 83.01\% & 85.77\% & 87\% & 96.87\% & 97.84\% & 90.43\% & 95\% & 92.9\% \\
 & Resnet101 + $\mathcal{L}_{style}$ & 96.54\% & 98.76\% & 87.77\% & 94\% & 95.54\% & 93.54\% & 90.7\% & 93\% & 92.49\% & 83.94\% & 91.99\% & 89\% & 95.7\% & 95.84\% & 91.98\% & 95\% & 92.9\% \\
 & Ours + $\mathcal{L}_{style}$ & 96.5\% & 98.4\% & 86.62\% & 94\% & 93.33\% & 92.52\% & 96.32\% & 94\% & 89.09\% & 79.48\% & 85.52\% & 85\% & 97.61\% & 96.13\% & 91.51\% & 95\% & 91.9\% \\
 & Ours + $\mathcal{L}_{style}$ + $\mathcal{L}_{attr}$ & 97.74\% & 98.66\% & 87.64\% & 95\% & 93.88\% & 92.39\% & 95.66\% & 94\% & 88.2\% & 79.41\% & 83.96\% & 84\% & 95.58\% & 95.59\% & 91.46\% & 94\% & 91.7\% \\
 & Ours + $\mathcal{L}_{style}$ + $\mathcal{L}_{attr}$ Balanced & 95.35\% & 98.24\% & 87.43\% & 94\% & 94.9\% & 95.46\% & 96.77\% & 96\% & 90.03\% & 82.2\% & 86.62\% & 86\% & 94.67\% & 95.9\% & 92.81\% & 94\% & 92.5\% \\
         \bottomrule
    \end{tabular}
    }
    \label{tab:fairness_ablations}
\end{table*}

\section{Additional Results}
Additional results of attribute prediction accuracy and fairness for different baselines and our models are reported in \autoref{tab:ablations}, and \autoref{tab:fairness_ablations} respectively.

    
\section{Taxonomy Description}
\label{sec:tax-details}
Here we present a complete textual description of the proposed hairstyle taxonomy.
\subsection{Scalp Regions}

We divide the head into 8 commonly understood regions to enable the granular labeling of attributes. 

\begin{itemize}
    \item Front
    \item Top
    \item Crown
    \item Nape
    \item Right Side
    \item Right Temple
    \item Left Side
    \item Left Temple
\end{itemize}

\subsection{Global Attributes}
Attributes that apply to the whole style.

\subsubsection{Bangs Style}
Bangs style indicates the shape of the bangs/fringe, or than none exists (i.e. the hair is gathered away from the forehead).
\begin{itemize}
    \item None
    \item Straight
    \item V-shaped
    \item U-shaped
    \item Inverted V-shaped
    \item Inverted U-shaped
    \item Diagonal top right to bottom left
    \item Diagonal top left to bottom right
    \item Other
\end{itemize}

\subsubsection{Bangs Length}
This is an alternative measure of length defining hair length relative to the brows used when bangs are present.
\begin{itemize}
    \item Above eyebrows (~<10cm)
    \item To eyebrows (~10cm)
    \item Below eyebrows (~>10cm)
\end{itemize}

\subsubsection{Hair Accessories}
This captures the presence of accessories in the hair style
\begin{itemize}
    \item None
    \item Headband
    \item Ribbons
    \item Hairnet
    \item Comb(s)
    \item Clip(s)
    \item Bead(s)
\end{itemize}

\subsubsection{Parting Location}
Location on the head where the hair parts/changes direction causing a visible valley.
\begin{itemize}
    \item Central
    \item Right side
    \item Left side
    \item Diagonal
    \item Zigzag
    \item Other
    \item None
\end{itemize}

\subsubsection{Hairline Shape}
Shape of the hairline, shape of the boundary line between the forehead and hair roots. We have listed the common lines and categories to handle other hairlines.
\begin{itemize}
    \item Straight
    \item Bell-shaped
    \item Receding/M-shaped
    \item Widow's peak
    \item Uneven/other
    \item I don't know 
\end{itemize}

\subsubsection{Hairline Position}
The position of the hairline relative to the forehead.
\begin{itemize}
    \item High
    \item Medium
    \item Low
    \item I don't know 
\end{itemize}

\subsubsection{Hairline Visibility}
This captures how much of the hairline was visible when specifying the hairline type and position.
\begin{itemize}
    \item Full
    \item Partially visible (left)
    \item Partially visible (right)
    \item Not visible
\end{itemize}

\subsubsection{Surface Appearance}
This describes the hair look (sheen, wetness/oil).
\begin{itemize}
    \item Matte
    \item Shiny
    \item Very shiny (oiled)
    \item Wet look
\end{itemize}

\subsubsection{Baby Hair}
This is hair around the forehead/hairline, particularly prominent in some coily hair styles.
\begin{itemize}
    \item No baby hair
    \item Unstyled
    \item Styled
    \item I don't know
\end{itemize}

\subsubsection{Hair Attribute Varies}
This indicates at least one hair attribute changes throughout hair length for a segment - a useful flag to indicate our taxonomy is not able to capture this hair style well.
\begin{itemize}
    \item No
    \item Yes
\end{itemize}

\subsection{Regional Attributes}
Attributes that can be specified at scalp region granularity.

\subsubsection{Hair type}
We adopt the Walker \cite{andrewalker} system though use names rather than numbers to avoid a sorting bias \cite{counteringracial}.
\begin{itemize}
    \item Coily (\citet{andrewalker} Type IV, kinky, afro-texture)
    \item Curly (\citet{andrewalker} Type III)
    \item Wavy (\citet{andrewalker} Type II)
    \item Straight (\citet{andrewalker} Type I)
\end{itemize}

\subsubsection{Strand Styling}
Strand styling describes how individual hair strands are combined, if at all.
\begin{itemize}
    \item None
    \item Other
    \item Twists/Ringlets
    \item Dreadlocks
    \item Braids
\end{itemize}

\subsubsection{Strand Thickness}
This attribute specifies the thickness of styled strands, when strand styling is not none.
\begin{itemize}
    \item Large (>2cm)
    \item Medium (1-2cm)
    \item Micro (<1cm)
\end{itemize}

\subsubsection{Hair Gathered}
Hair gathered indicates a hair gathering is present at the region specified, all other regions should be not gathered - even if strands are flowing into a gathering. For example, a subject with a pony tail only might have all regions marked "not gathered" except for the crown which might be "pony tail single". Values containing "multiple" indicates there are multiple gatherings for this region, not that there are multiple gatherings globally.
\begin{itemize}
    \item None, not gathered
    \item Tucked behind the ear
    \item Bun, single
    \item Bun, multiple
    \item Pony tail, single
    \item Pony tail, multiple
    \item Attached to the skin (cornrows, French plaits)
    \item Knot, single
    \item Knot, multiple
    \item Gathered, other, not listed
    \item Gathered, gathering style not visible
\end{itemize}

\subsubsection{Hair Direction}
Hair direction indicates the direction strands are pointing for a region relative to the direction of the subjects' face.
\begin{itemize}
    \item Brushed/flowing down
    \item Brushed/swept to the side
    \item Brushed/gathered up
    \item Pointing out
\end{itemize}

\subsubsection{Hair Length}
Hair length applies to all regions except front in presence of bangs, where bangs length should be used instead. Shorter lengths are measured in centimeters, longer lengths are measured relative to features of the person for labeling practically. Common clipper guard sizes are listed.
\begin{itemize}
    \item No hair/Bald (clipper 0)
    \item Shaved, roots visible (clipper 0.5)
    \item Very short (<1cm, clipper 1-3)
    \item Short (1-5cm, clipper 4-10)
    \item Ear length
    \item Chin length
    \item Shoulder length
    \item Armpit length
    \item Mid-back length
    \item Waist length or longer
    \item Hair not visible
\end{itemize}

\subsubsection{Layering}
Layering captures how hair length varies within a region, often a distinctive attribute for hair styles - affecting how hair appears and sits.
\begin{itemize}
    \item None/Single length
    \item Textured/Layered
    \item Taper
    \item Fade
\end{itemize}

\subsubsection{Decorative patterns}
Indicates the presence of patterns cut into the hair. These can be co-located with other features, for example, a line of no hair cut out along the parting.
\begin{itemize}
    \item None
    \item Decorated
\end{itemize}

\bibliographystyle{ACM-Reference-Format}
\bibliography{refs}

\end{document}